\documentclass[conference]{IEEEtran}
\IEEEoverridecommandlockouts

\usepackage{cite}
\usepackage{amsmath,amssymb,amsfonts}
\usepackage{graphicx}
\usepackage{textcomp}
\usepackage{xcolor}
\usepackage{url}
\usepackage{balance}
\usepackage{placeins}
\usepackage{array}

\usepackage{eso-pic}
\AddToShipoutPictureBG*{%
  \AtPageLowerLeft{%
    \makebox[\paperwidth]{%
      \raisebox{0.32in}{%
        \parbox{0.84\paperwidth}{\centering\scriptsize
          \copyright~2026 IEEE. Personal use of this material is permitted. Permission from IEEE must be obtained for all other uses, in any current or future media, including reprinting/republishing this material for advertising or promotional purposes, creating new collective works, for resale or redistribution to servers or lists, or reuse of any copyrighted component of this work in other works.\\[2pt]
          Accepted for publication in the 2026 IEEE 8th International Conference on Cognitive Machine Intelligence (CogMI).}}}}}
          
\begin{document}

\title{Loud Failures, Quiet Failures: Fault Detection and\\Recovery in Tool-Using Language Model Agents}

\author{\IEEEauthorblockN{Obada Kraishan}
\IEEEauthorblockA{College of Media and Communication \\
Texas Tech University \\
Lubbock, TX, USA \\
omareikr@ttu.edu}}

\maketitle

\begin{abstract}
Tool-using agents are usually scored on whether they finish a task while the tools work. Deployments are less forgiving: services time out, endpoints disappear, parameter names change, and results come back well formed but wrong. We ask what agents do when the tool layer misbehaves. Prior work has shown that language models over-trust tool outputs that fail silently; we ask how that over-trust plays out across the stages of failure handling in multi-turn agents. Wrapping the executable environments of an established function-calling benchmark in a fault-injection layer, we inject one of four typed faults at a controlled point in the trajectory and record whether the agent notices, changes plan, recovers the task, or repeats itself. Six models from three families, half of them reasoning variants, ran 1{,}920 trials over 24 multi-step tasks. Agents treat a failure as a problem in 91.3\% of trials when the tool returns an explicit error, but in 58.8\% of trials when the tool returns a plausible wrong value, against a 26.8\% rate of reporting problems when nothing was wrong ($h = 0.80$ between the two fault kinds). Reasoning models are not better placed here: paired against instruct siblings from the same family, they notice less ($-9.3$ points, $p < .001$) and change plan more ($+10.4$ points, $p < .001$), and the two effects leave recovery unchanged ($p = .512$ in a task-clustered model). Because agents are stochastic, two fault-free runs of the same task end in the same state only 63.3\% of the time; against that baseline, only a missing tool clearly lowers recovery (39.9\%), while timeouts, schema drift, and corruption stay within run-to-run variation. After a fault, agents return to the same tool three or more times in a row in up to 22.2\% of trials, though strictly identical repeats are rare. Adding a line to the prompt asking the agent to check each result did not move detection. The pattern points at a specific weakness: agents respond to the error channel rather than to the content of what a tool returns, so failures that stay inside the expected format pass through.
\end{abstract}

\begin{IEEEkeywords}
tool-using agents, fault injection, failure detection, agent evaluation, cognitive flexibility, reasoning models
\end{IEEEkeywords}

\section{Introduction}

When a language model calls tools, its mistakes no longer stay on the page. An agent books, transfers, writes, and deletes, and the record of what it did is the tool-call log rather than the prose around it. Benchmarks for these systems have grown quickly, and they measure something reasonable: given a working environment, does the agent reach the goal \cite{patil2024gorilla, qin2024toolllm, liu2024agentbench}. That question leaves out the part of deployment that operators actually worry about. Services time out. Endpoints are retired. A field gets renamed in a minor release. A cache returns a stale value that is correctly typed and quietly wrong. Evaluations that add this stress show how much a clean run hides: success measured once on working tools falls when the same tasks are repeated \cite{yao2024taubench} or when faults are injected into the tool layer \cite{gupta2026reliabilitybench}, and models tend to trust tool outputs that are wrong but carry no error signal \cite{sun2024tools, zhu2026toolmaze}.

Handling those events is not a single skill. An agent must first register that something is off, then decide on a different course, then still reach the goal, and it must do all of this without treating ordinary friction as catastrophe. Those components separate cleanly in the human literature on monitoring and control of automation, where the failure to notice a malfunctioning system is studied apart from the failure to respond to it \cite{parasuraman1997humans, endsley1995measurement}, and where repeating an action that has stopped working is treated as its own phenomenon rather than as generic incompetence \cite{miyake2000unity}. Agent evaluations rarely make the same separations. A trial that ends without the goal is scored as a failure whether the agent never saw the problem or saw it and could not route around it, and those two cases call for different fixes.

This paper measures the components separately. We take the multi-turn split of the Berkeley Function-Calling Leaderboard, whose environments execute locally as stateful Python objects, and wrap tool execution in a layer that injects one typed fault at a controlled point in the trajectory. Four faults are used: a timeout, a tool that disappears from the registry, a renamed parameter, and a returned value that is corrupted while keeping its shape. A no-fault condition provides the baseline. For every trial we record whether the agent treated the result as a problem, whether its next action differed from the failed one, whether the environment ended in the state the same model reaches on an independent clean run, and whether it called the same tool three or more times in a row.

The contrast that organises the results is not the name of the fault but its visibility. Three of the four faults surface an explicit error; the agent has only to read it. Corruption does not: the call succeeds, the payload is well formed, and only the content is wrong. That silent errors are harder to catch than explicit ones has been reported before, for single tool outputs judged on request \cite{sun2024tools} and for recovery in tasks with alternative tool paths \cite{zhu2026toolmaze}. What those studies leave open is how the stages of failure handling relate to one another within the same tasks and models: whether an agent that misses a failure also fails to recover, whether an agent that acts more also notices more, and how often agents report problems when nothing is wrong.

The study is organised around three research questions, with a fourth, secondary analysis:
\begin{itemize}
\item \textbf{RQ1.} Does an agent's detection of a tool failure depend on whether the failure surfaces an explicit error?
\item \textbf{RQ2.} Do reasoning models handle tool failures better than matched instruct siblings from the same family?
\item \textbf{RQ3.} What does a single tool failure cost in recovery, repeated calls, and effort?
\end{itemize}
The secondary analysis asks whether an instruction to check each tool result changes detection.

Our contributions are:

\begin{itemize}
\item a fault-injection layer for an existing executable agent benchmark, with typed faults that fire once at a controlled step and leave the environment consistent, released with the frozen task suite and the analysis code;
\item a measurement scheme that separates noticing, replanning, recovery, and repetition, with recovery scored against the state the same model reaches without a fault, and judged against the agreement of two fault-free runs, so that neither baseline competence nor run-to-run variation is mistaken for a fault effect;
\item evidence from 1{,}920 trials in multi-turn executable environments that extends earlier single-step findings on silent errors: detection follows the error channel against a no-fault baseline, and reasoning variants notice less than their instruct siblings while acting more, with no net gain in recovery.
\end{itemize}

We release the harness, the frozen suite, the full trajectories, the scored dataset, and the analysis scripts.\footnote{\url{https://github.com/obadaKraishan/brittle-agents}}

\section{Related Work}

Four lines of work meet in this study: how tool-using agents are built and scored, how faults in the tool layer have been studied, what reasoning variants of those models do, and how failure handling has been treated in cognitive and engineering research.

\subsection{Tool use and agent evaluation}

Interleaving reasoning with tool calls became standard practice after ReAct \cite{yao2023react}, and models are now trained to call functions directly \cite{schick2023toolformer, patil2024gorilla}. Benchmarks followed. ToolLLM covers large API collections \cite{qin2024toolllm}; AgentBench spans several interactive environments \cite{liu2024agentbench}; WebArena places agents in realistic web tasks \cite{zhou2024webarena}; $\tau$-bench adds users and domain rules to conversational tool use \cite{yao2024taubench}. The Berkeley Function-Calling Leaderboard, which we build on, is unusual in that its multi-turn split ships executable stateful environments rather than reference call strings, so an agent's effect on the world can be checked directly \cite{patil2024gorilla}.

Recent critiques of this literature focus on what the scores hide. Kapoor et al.\ argue that agent benchmarks reward narrow accuracy at the expense of cost and reliability, and that held-out evaluation practice is weaker than in ordinary machine learning \cite{kapoor2024agents}. Cemri et al.\ build a taxonomy of failure modes for multi-agent systems and find that most failures come from specification and coordination rather than from the base model's limits \cite{cemri2025why}. Both point toward measuring the shape of failures rather than counting them, which is what we try to do here for the single-agent case.

\subsection{Faults in tool-using agents}

A small set of studies injects failures into the tool layer itself, and they are the closest work to ours. Sun et al.\ showed that language models over-trust tools that fail silently: with a deliberately faulty calculator and in an embodied planning setting, models often accepted incorrect outputs that carried no error signal, and a short in-context disclaimer improved detection \cite{sun2024tools}. PALADIN injects execution errors from an existing failure taxonomy into ToolBench trajectories and trains agents to recover, raising recovery from 23.75\% to 89.86\% against untrained agents \cite{vuddanti2025paladin}. ReliabilityBench adapts chaos engineering to agents, injecting timeouts, rate limits, partial responses, and schema drift across four domains and scoring success by end-state equivalence \cite{gupta2026reliabilitybench}. ToolMaze crosses explicit and implicit perturbations with transient and permanent ones in tasks with alternative tool paths, and finds that recovery falls most under implicit failures, which it traces to over-trust in corrupted outputs \cite{zhu2026toolmaze}.

Table~\ref{tab:related} sets these studies beside ours. Their numbers are not directly comparable to ours, since tasks, faults, and outcome definitions differ, but their agreement on silent failures is itself informative, and our results add to it. Our study shares the fault-injection approach and the observation that silent failures are the hardest. It differs in what it measures: detection is coded separately from replanning and recovery and given a no-fault baseline, recovery is scored against each model's own clean run, and reasoning and instruct models are compared within the same family.

\begin{table}[!t]
\centering
\caption{Studies that inject faults into the tool layer, by what they measure.}
\label{tab:related}
\footnotesize
\setlength{\tabcolsep}{2.5pt}
\begin{tabular}{>{\raggedright\arraybackslash}p{1.6cm}>{\raggedright\arraybackslash}p{1.55cm}>{\raggedright\arraybackslash}p{2.05cm}>{\raggedright\arraybackslash}p{1.95cm}}
\hline
Study & Setting & Faults & Primary outcome \\
\hline
Sun et al.\ \cite{sun2024tools} & calculator; embodied planning & silent tool errors & detection when asked to accept or reject \\
PALADIN \cite{vuddanti2025paladin} & ToolBench & execution errors & recovery rate after training \\
Reliability\-Bench \cite{gupta2026reliabilitybench} & four simulated domains & timeouts, rate limits, partial responses, schema drift & success under fault, by end state \\
ToolMaze \cite{zhu2026toolmaze} & multi-path tool tasks & explicit/implicit, transient/permanent & recovery rate under perturbation \\
This work & BFCL multi-turn & timeout, missing tool, schema drift, silent corruption & detection, replanning, recovery, repetition, separately \\
\hline
\end{tabular}
\end{table}

\subsection{Reasoning models}

Models that spend a controllable budget on an internal deliberation phase before answering now sit alongside their instruction-tuned siblings \cite{jaech2024openai, guo2025deepseek, yang2025qwen3}, and extra test-time computation buys real gains on tasks with verifiable answers \cite{wei2022chain, snell2024scaling}. Whether the deliberation is what produces those gains is contested. Turpin et al.\ showed that chain-of-thought text can rationalise an answer driven by a cue the text never mentions \cite{turpin2023language}; Chen et al.\ found the same inside reasoning models, which name the hints that moved them only part of the time \cite{chen2025reasoning}; Shojaee et al.\ report reasoning effort that collapses as problems get harder \cite{shojaee2025illusion}. Our design gives these models a different kind of test. Nothing in a fault trial rewards longer deliberation by itself, and the question is whether the extra phase helps an agent notice that its environment has stopped cooperating.

\subsection{Failure handling}

Injecting faults to see how a system copes is old practice in dependable computing \cite{hsueh1997fault} and current practice in production engineering, where controlled failures are introduced into live services to expose weaknesses before customers find them \cite{basiri2016chaos}. What that tradition measures is the system's recovery, not its awareness, because conventional software has no awareness to measure. Agents do, in the limited sense that their output states a reading of the situation, and that opens the door to the distinction the human factors literature has long drawn between detecting an automation failure and responding to it \cite{parasuraman1997humans, mosier1999automation}. Work on self-correction in language models sits on the response side of that line: Reflexion has agents write lessons for themselves after a failed attempt \cite{shinn2023reflexion}, and Self-Refine loops critique and revision \cite{madaan2023selfrefine}. Both assume the problem has been identified. The present study asks how often that assumption holds.

\section{Method}

This section describes the tasks, the faults and how they were injected, the agent protocol, the measures, and the checks we ran on the resulting data.

\subsection{Tasks and environments}

Tasks come from the multi-turn split of the Berkeley Function-Calling Leaderboard, which pairs each scenario with Python classes that hold state and expose tools as methods \cite{patil2024gorilla}. We drove those classes directly rather than through the official evaluation harness, which is what makes controlled injection possible: the tool layer is ours, so a fault can be placed at a chosen step and nowhere else.

Of 400 candidate tasks, 312 passed three filters: every environment class they use imports and runs locally, the task has between three and ten user turns, and an executable ground-truth call sequence is available. From those we sampled 24 tasks with a fixed seed, balanced across environment domains, and froze the suite before any trial was run. The frozen set spans 16 domain combinations, including a file system, a trading interface, travel booking, vehicle control, messaging, and a ticketing system, with a median of five user turns per task.

\subsection{Fault taxonomy}

Four faults were chosen to separate failures an agent can read from failures it must infer, with a fifth no-fault condition as baseline (Table~\ref{tab:faults}). \textit{Timeout} raises a timeout error. \textit{Missing tool} removes a tool from the registry and reports it as unavailable. \textit{Schema drift} rejects the call with an unknown-argument error, as if a parameter had been renamed. \textit{Silent corruption} lets the call succeed and returns a value of the same type and shape as the real one, altered so that its content is wrong: a number moved by 20 to 60 percent, a list with an element dropped, a field replaced. When the real result is empty, a placeholder string is returned instead; this happened in 2 of the 374 analysed corruption trials. The first three place an error string in front of the agent. The fourth does not, and we refer to these two situations as loud and quiet failures.

\subsection{Injection procedure}

The fault fires once, on the first eligible call at or after step $\lfloor T/2 \rfloor$, where $T$ is the number of user turns in the task. In practice this placed faults early: the median firing step was 3, against a median of 7 calls in the reference solutions, and 42.1\% of faults fired at step 1 or 2. Timing differs by fault type, because the two faults that target a specific tool wait for the agent to call it: timeouts and corruption fired at a median of step 2, missing tool and schema drift at step 5. Later calls behave normally, which means recovery is available by construction rather than blocked by the manipulation.

Two properties of the layer matter for interpretation. Faults do not leave the environment half-changed: corruption alters the returned value only, and the underlying state is what a successful call would have produced. Second, the two faults that must target a particular tool (missing tool and schema drift) take that tool from the task's ground-truth call path rather than from a heuristic, because a fault attached to a tool the agent never calls is not a fault at all; even so, these two faults fired in 71\% and 74\% of their trials. Timeout fired in 99\% of its trials and corruption in 97\%, since neither depends on a specific tool. Trials in which the fault never fired are reported in Table~\ref{tab:faults} and excluded from analysis, leaving 1{,}694 of 1{,}920 trials.

A self-test accompanies the layer and is run before collection. It asserts, on every task, that the clean condition never faults, that each fault type fires exactly once and at or after its target step, that corruption changes the returned value while preserving its type (apart from the empty-result case above) and leaving environment state untouched, and that the same inputs produce the same fault behaviour on repeated runs.

\begin{table}[!t]
\centering
\caption{Conditions, visibility, and how often each fault fired.}
\label{tab:faults}
\footnotesize
\setlength{\tabcolsep}{4pt}
\begin{tabular}{llccc}
\hline
Condition & Visibility & Trials & Fired & Analysed \\
\hline
clean & baseline & 384 & -- & 384 \\
timeout & loud & 384 & 99\% & 379 \\
missing tool & loud & 384 & 71\% & 273 \\
schema drift & loud & 384 & 74\% & 284 \\
silent corruption & quiet & 384 & 97\% & 374 \\
\hline
\end{tabular}
\par\smallskip\scriptsize Loud faults surface an explicit error; the quiet fault returns a well-formed but incorrect value. Missing tool and schema drift fire only when the agent calls the affected tool.
\end{table}

\subsection{Agent protocol}

Each trial follows the benchmark's multi-turn structure. A user turn is appended to the conversation, the model acts until it replies without calling a tool, and the next turn follows. Tools are exposed as JSON schemas through the provider's native function-calling interface, and every call passes through the injection layer. The system prompt tells the agent to work step by step, to call one tool at a time, and to keep working toward the goal if a tool behaves unexpectedly; it does not mention the experiment. Trials stop after 15 tool calls, a bound we report rather than hide, since exhausting it is one of the outcomes of interest.

Six models were used, forming three matched pairs so that the reasoning contrast never crosses model families (Table~\ref{tab:design}). Reasoning variants ran with a 2{,}048-token thinking budget. All calls were routed through a single gateway to keep the interface uniform, sampling temperature was 1.0, and each cell was run twice; the provider's sampling is not seeded, so the two runs are independent samples. The full design gives $24 \times 5 \times 2 = 240$ trials per model, plus a second arm for one matched pair described in Section~\ref{sec:scaffold}, for 1{,}920 trials in total. A seventh and eighth model, a free-tier Nemotron pair, were dropped before analysis after repeated rate-limit failures, with no completed trials; among the six models analysed, no trial failed at the API level.

\begin{table}[t]
\centering
\caption{Model panel. Each family contributes a matched reasoning and instruct pair.}
\label{tab:design}
\footnotesize
\setlength{\tabcolsep}{4pt}
\begin{tabular}{lllcc}
\hline
Model & Family & Type & Trials & Prompt arm \\
\hline
claude-haiku-4.5-thinking & claude & reasoning & 480 & yes \\
claude-haiku-4.5 & claude & instruct & 480 & yes \\
deepseek-r1 & deepseek & reasoning & 240 & -- \\
deepseek-v3 & deepseek & instruct & 240 & -- \\
qwen3-thinking & qwen & reasoning & 240 & -- \\
qwen3-instruct & qwen & instruct & 240 & -- \\
\hline
\end{tabular}
\par\smallskip\scriptsize 1{,}920 trials over 24 tasks, five conditions, two seeds. The secondary scaffolding analysis doubles one pair.
\end{table}

\subsection{Measures}

Let $S_a$ be the set of public state fields of the environment at the end of a trial and $S_r$ the same fields from a reference run. State agreement is the share of reference fields the trial reproduces:

\begin{equation}
A(S_a, S_r) \;=\; \frac{\left|\{f \in S_r : S_a[f] = S_r[f]\}\right|}{|S_r|} \;\in\; [0, 1].
\label{eq:agreement}
\end{equation}

Recovery uses each model as its own reference. Writing $S_c^{m,t,s}$ for the end state model $m$ reaches on task $t$ with seed $s$ in the clean condition,

\begin{equation}
\text{recovered} \;=\; \mathbb{1}\!\left[\, A\!\left(S_a,\; S_c^{m,t,s}\right) = 1 \,\right].
\label{eq:recovery}
\end{equation}

Scoring against the model's own clean run rather than against the benchmark's ground truth removes baseline competence from the comparison: a model that cannot finish a task without a fault is not counted as failing to recover from one. Because sampling is not seeded, two clean runs of the same model and task need not end in the same state. Clean trials are therefore scored the same way: each clean run without the prompt line is scored against the other repetition of the same model and task, and each clean run in the prompt arm (Section~\ref{sec:scaffold}) against the run without the line. Pooled, these give the run-to-run agreement that fault trials are compared with (58.3\% and 78.1\% for the two groups, 63.3\% overall). Effort is measured on the same footing, as steps used relative to the clean runs of that model and task:

\begin{equation}
\text{cost} \;=\; n_{\text{steps}} \;-\; \frac{1}{|C|}\sum_{c \in C} n_{\text{steps}}^{\,c},
\qquad C = C^{m,t}_{\text{clean}}.
\label{eq:cost}
\end{equation}

Three further measures come from the trajectory. \textit{Replanning} is true when the first action after the fault differs from the action that failed, by tool or by arguments. \textit{Repetition} is true when three or more consecutive calls after the fault go to the same tool, whether or not the arguments change; strictly identical repetition, with the same arguments, is reported alongside. Neither is defined for clean trials. \textit{Budget exhaustion} is true when the trial ends at the step cap.

\subsection{Coding failure detection}

Whether an agent treated a result as a problem is a judgement about language, so it was coded by a model and validated against a human. The judge sees one tool call, the result it returned, and what the agent said next. It never learns whether a fault was injected or which condition the trial belongs to, and clean trials are coded the same way using the trajectory midpoint, which gives a baseline for how often an agent reports a problem when there is none.

We first tried a three-way scheme separating no reaction, acknowledgement, and cases where the agent noticed an anomaly and explained it away. One author hand-coded 100 trials, drawn at random from the analysable set during an earlier scoring pass, blind to the judge's labels; the coded sheet is released with the data. The three-way scheme reached only $\kappa = .47$; most disagreements (19 of 32) were trials the human read as acknowledgement and the judge labelled as explaining away. On the 92 trials that both the author and the judge coded, collapsing to the binary distinction (did the agent treat this result as a problem) gave 85.9\% agreement and $\kappa = .69$, which is the measure used throughout. The three-way labels are retained in the released data but no claim rests on them.

\subsection{Statistical analysis}
Trials are not independent: the same 24 tasks recur across models, faults, and seeds. Inference therefore rests on task-clustered estimates, namely binomial generalized estimating equations with an exchangeable working correlation clustered on task, and bootstrap intervals that resample tasks. Per-family contrasts in RQ2 use task-clustered models on the matched pairs, which balance fault type by design. Two-proportion $z$-tests, and paired $t$-tests in the secondary analysis, are reported alongside as descriptive comparisons. Holm correction is applied within each family of tests, and conclusions rest on the clustered results.

\subsection{Data quality}
\label{sec:dataquality}

Three facts about the dataset belong with the results. Detection on clean trials whose target call succeeded ran at 26.8\%, and this rate is treated as the baseline rather than as noise: these environments produce genuine errors of their own, such as authentication and balance messages, and clean trials whose target call returned one of those were acknowledged 90\% of the time, which is correct behaviour rather than a false alarm. Trials ended at the step cap 27\% of the time overall. One model, DeepSeek-R1, wrote tool-call syntax into its message text rather than emitting a structured call in 1.8 turns per trial on average; those calls never execute, and the behaviour is a property of that model in this interface that we note as a limitation.

\section{Results}

This section answers the three research questions in turn and then reports the secondary analysis on prompting the agent to check its results.

\subsection{RQ1: Detection depends on visibility, not severity}

\begin{figure}[!t]
\centering
\includegraphics[width=\columnwidth]{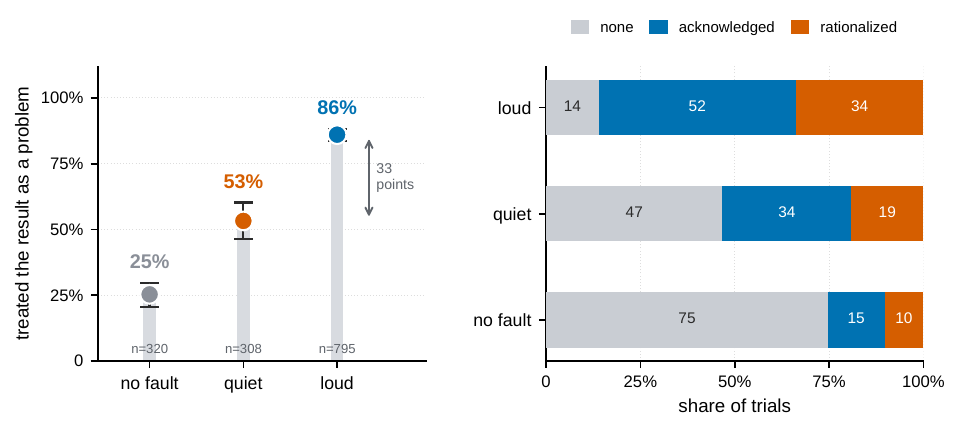}
\caption{Detection by failure visibility. Left: share of trials in which the agent treated the tool result as a problem, with task-clustered bootstrap intervals. Right: how those responses divide, where ``none'' means the result was used or reported as if it were fine.}
\label{fig:detection}
\end{figure}

Figure~\ref{fig:detection} and Table~\ref{tab:rq1} give the first result. When a tool returned an explicit error, agents treated it as a problem in 91.3\% of trials (95\% CI [88.3, 94.1], $n = 936$). When a tool returned a plausible wrong value, they did so in 58.8\% of trials ([45.7, 71.2], $n = 374$). The gap is 32.5 points, $z = 13.86$, $p < .001$, $h = 0.80$. Both sit above the rate at which agents reported a problem when the target call had in fact succeeded, 26.8\% ([17.8, 36.5], $n = 306$); quiet failures clear that baseline by 32.0 points ($z = 8.36$, $p < .001$, $h = 0.66$) and loud ones by 64.5 points ($z = 22.77$, $p < .001$, $h = 1.46$).

The cleanest version of this contrast is between timeouts and corruption, which fired at the same median step: 91.6\% against 58.8\%. The three loud faults land within a point of each other---91.6\% for timeouts, 91.2\% for a missing tool, 91.2\% for a renamed parameter---despite differing in what the agent must do next. Corruption sits 32 points below all three. A binomial generalized estimating equation with an exchangeable working correlation clustered on task puts quiet failures at $\mathrm{OR} = 3.96$ against the no-fault baseline ($z = 6.48$, $p < .001$) and loud failures at $\mathrm{OR} = 28.61$ ($z = 12.14$, $p < .001$).

The right panel of Figure~\ref{fig:detection} shows the composition behind those rates. Under a loud fault, 4\% of trials pass without the agent treating the result as a problem. Under corruption that figure is 39\%. In those trials the agent reads a wrong value and carries it forward, sometimes reconciling it with what the conversation had established earlier.

\begin{table}[!t]
\centering
\caption{Detection rates with task-clustered bootstrap intervals.}
\label{tab:rq1}
\footnotesize
\setlength{\tabcolsep}{4pt}
\begin{tabular}{lccc}
\hline
Condition & $n$ & Detected & 95\% CI \\
\hline
no fault (baseline) & 306 & 26.8\% & [17.8, 36.5] \\
quiet failure & 374 & 58.8\% & [45.7, 71.2] \\
loud failure & 936 & 91.3\% & [88.3, 94.1] \\
\hline
timeout & 379 & 91.6\% & [88.0, 94.5] \\
missing tool & 273 & 91.2\% & [86.5, 95.2] \\
schema drift & 284 & 91.2\% & [86.3, 95.4] \\
silent corruption & 374 & 58.8\% & [45.7, 71.2] \\
\hline
\end{tabular}
\par\smallskip\scriptsize All fault conditions differ from the baseline at $p < .001$ after Holm correction.
\end{table}

\subsection{RQ2: Reasoning models act more and notice less}

\begin{figure}[!t]
\centering
\includegraphics[width=\columnwidth]{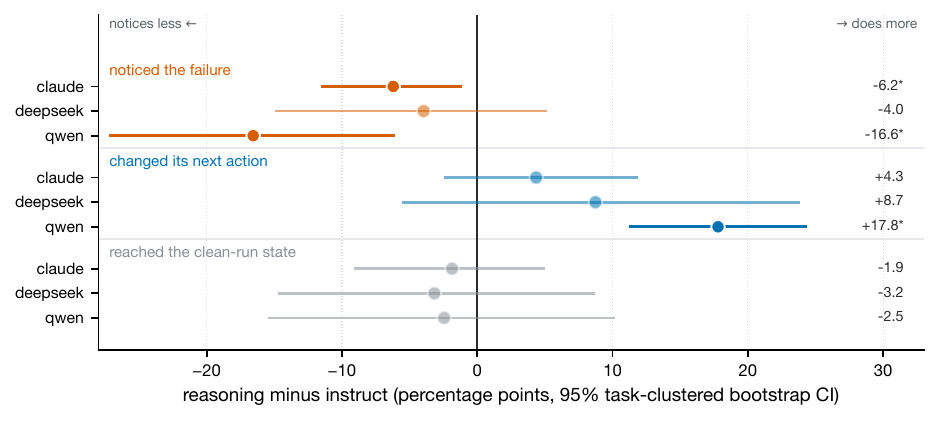}
\caption{Reasoning variants minus matched instruct siblings on fault trials, paired within family by task, fault, and seed. Points are mean differences in percentage points with task-clustered 95\% bootstrap intervals; asterisks mark contrasts significant in task-clustered models after Holm correction. Noticing sits left of zero in every family while acting sits right of it, and recovery straddles the line.}
\label{fig:reasoning}
\end{figure}

Pairing each reasoning model against the instruct sibling from its own family, on the same task, fault, and seed, produces the pattern in Figure~\ref{fig:reasoning} and Table~\ref{tab:rq2}. On detection the reasoning model is lower in all three families: $-6.2$ points for Claude ($\mathrm{OR} = 0.61$, $p_{\text{Holm}} = .042$), $-4.0$ points for DeepSeek ($p_{\text{Holm}} = .434$), and $-16.6$ points for Qwen ($\mathrm{OR} = 0.40$, $p_{\text{Holm}} = .003$), in task-clustered models on the matched pairs. Pooled over 450 pairs the difference is $-9.3$ points (task-clustered 95\% CI $[-14.3, -5.3]$), and a task-clustered model with fault type as a covariate gives $\mathrm{OR} = 0.56$ for the reasoning term ($z = -4.28$, $p < .001$). Qwen carries the clearest single-family result. Claude's difference holds on the matched pairs but is not robust to specification: modelled over all of its fault trials with fault type as a covariate, it is not significant ($p_{\text{Holm}} = .209$). DeepSeek, whose reasoning model sometimes wrote tool calls as text (Section~\ref{sec:dataquality}), shows the smallest difference.

Replanning runs the other way. Reasoning models change their next action more often, by 4.3 points for Claude, 8.7 for DeepSeek, and 17.8 for Qwen, pooling to $+10.4$ points (task-clustered 95\% CI $[+5.6, +15.7]$; $\mathrm{OR} = 2.02$, $z = 3.52$, $p < .001$). Of the single families, only Qwen's difference is reliable ($\mathrm{OR} = 4.23$, $p_{\text{Holm}} < .001$; Claude and DeepSeek $p_{\text{Holm}} = .417$).

Recovery shows neither advantage. No family differs ($p_{\text{Holm}} = 1.000$ throughout), and the pooled estimate is $-2.4$ points (task-clustered 95\% CI $[-9.3, +4.5]$; $\mathrm{OR} = 0.91$, $p = .512$). Whatever the extra deliberation phase changes about how these agents behave after a fault, it does not change how often they end up where they were going.

\begin{table*}[!t]
\centering
\caption{Reasoning versus instruct siblings on fault trials, paired within family by task, fault, and seed.}
\label{tab:rq2}
\footnotesize
\setlength{\tabcolsep}{5pt}
\begin{tabular}{llcccccc}
\hline
Outcome & Family & Reasoning & Instruct & $\Delta$ (points) & Clustered 95\% CI & OR & $p_{\text{Holm}}$ \\
\hline
detection & claude & 82.0\% & 88.2\% & $-6.2$ & $[-11.4, -1.2]$ & 0.61 & .042 \\
detection & deepseek & 78.6\% & 82.5\% & $-4.0$ & $[-14.8, +5.0]$ & 0.78 & .434 \\
detection & qwen & 67.5\% & 84.0\% & $-16.6$ & $[-27.2, -6.2]$ & 0.40 & .003 \\
replanning & claude & 75.2\% & 70.8\% & $+4.3$ & $[-2.3, +11.8]$ & 1.25 & .417 \\
replanning & deepseek & 74.6\% & 65.9\% & $+8.7$ & $[-5.5, +23.8]$ & 1.52 & .417 \\
replanning & qwen & 92.6\% & 74.8\% & $+17.8$ & $[+11.3, +24.2]$ & 4.23 & $<$.001 \\
recovery & claude & 66.5\% & 68.3\% & $-1.9$ & $[-9.0, +4.9]$ & 0.92 & 1.000 \\
recovery & deepseek & 38.9\% & 42.1\% & $-3.2$ & $[-14.6, +8.6]$ & 0.88 & 1.000 \\
recovery & qwen & 44.2\% & 46.6\% & $-2.5$ & $[-15.3, +10.1]$ & 0.91 & 1.000 \\
\hline
\end{tabular}
\par\smallskip\scriptsize Negative $\Delta$ means the reasoning variant scored lower. OR and $p_{\text{Holm}}$ from task-clustered models on the matched pairs, Holm-corrected across families. Pooled over 450 pairs: detection $-9.3$ points ($p < .001$), replanning $+10.4$ points ($p < .001$), recovery $-2.4$ points ($p = .512$).
\end{table*}

\subsection{RQ3: What faults cost}

Judging what a fault costs requires knowing what happens without one. Two fault-free runs of the same model and task end in the same state in 63.3\% of trials (task-clustered 95\% CI $[53.1, 73.2]$), and the figure varies widely by model, from 84.4\% for the Claude instruct model to 41.7\% for DeepSeek-R1, DeepSeek-V3, and the Qwen3 thinking model. Against that baseline, only a missing tool clearly lowers recovery: 39.9\% ($[28.1, 52.4]$, $h = -0.47$; $\mathrm{OR} = 0.34$, $p_{\text{Holm}} < .001$). Timeouts (59.6\%), schema drift (59.2\%), and corruption (60.4\%) stay within run-to-run variation ($\mathrm{OR}$ 0.80 to 0.88, $p_{\text{Holm}} \geq .117$); schema drift reaches significance only once model is added as a covariate ($\mathrm{OR} = 0.76$, $p_{\text{Holm}} = .024$). Losing a capability is the one fault no retry can undo, and it is the one that clearly moves end states.

That corruption leaves end states largely intact while going unnoticed in 41\% of trials is less reassuring than it sounds. Recovery here is agreement in environment state, and a wrong value that the agent reads rather than writes changes what it reports, not what it does to the environment.

\begin{figure}[!t]
\centering
\includegraphics[width=\columnwidth]{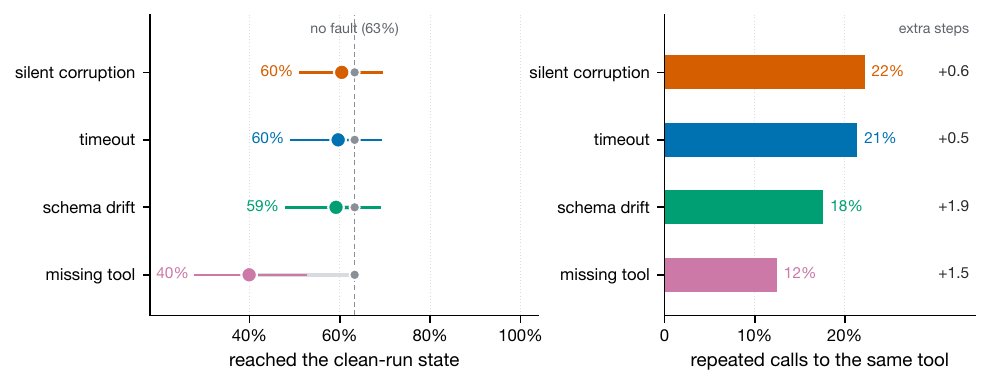}
\caption{Consequences of a single fault. Left: share of trials reaching the state of an independent fault-free run of the same model and task, against the agreement of two fault-free runs (63.3\%). Right: how often the agent called the same tool three or more times in a row after the fault, with the extra steps a fault costs relative to a clean run listed alongside.}
\label{fig:cost}
\end{figure}

Effort tells a second story (Figure~\ref{fig:cost}, Table~\ref{tab:rq3}). After a fault, agents called the same tool three or more times in a row in 12.5\% to 22.2\% of trials, most often under corruption (22.2\%) and timeouts (21.4\%). These are mostly retries with changed arguments rather than loops: strictly identical repetition, with the same tool and arguments, occurred in 0.8\% to 1.8\% of fault trials. Trials end at the step cap between 24.8\% and 31.7\% of the time under faults, against 23.2\% when nothing goes wrong. The extra steps a fault costs relative to a clean run are modest for timeouts ($+0.53$) and corruption ($+0.56$) and larger where the agent has to rebuild its plan, at $+1.47$ for a missing tool and $+1.85$ for a renamed parameter.

Whether noticing helps deserves a careful answer. Compared directly, trials in which the agent treated the result as a problem recovered at 55.6\% and trials in which it did not at 55.7\%, a difference of nothing at all ($z = -0.03$, $p = .974$). That comparison is confounded: corruption is both the least detected fault and one of the more recoverable ones, so pooling the conditions cancels the effect. Holding fault type fixed, noticing is associated with recovery ($\mathrm{OR} = 1.51$, $z = 2.91$, $p = .004$), but the association weakens once model is also held fixed ($\mathrm{OR} = 1.29$, $p = .098$), so part of it reflects which models notice more rather than what noticing does. Detection may be worth having; this study cannot show that it is sufficient.

\begin{table}[!t]
\centering
\caption{Recovery and effort by condition.}
\label{tab:rq3}
\footnotesize
\setlength{\tabcolsep}{3pt}
\begin{tabular}{lccccc}
\hline
Condition & Recovered & Same tool & Identical & Cap hit & Extra steps \\
\hline
clean & 63.3\% & -- & -- & 23.2\% & $+0.01$ \\
timeout & 59.6\% & 21.4\% & 1.1\% & 24.8\% & $+0.53$ \\
missing tool & 39.9\% & 12.5\% & 1.1\% & 29.3\% & $+1.47$ \\
schema drift & 59.2\% & 17.6\% & 1.8\% & 31.7\% & $+1.85$ \\
silent corruption & 60.4\% & 22.2\% & 0.8\% & 25.9\% & $+0.56$ \\
\hline
\end{tabular}
\par\smallskip\scriptsize Clean recovery is the agreement of two independent fault-free runs. ``Same tool'' and ``Identical'' are three or more consecutive calls after the fault, to the same tool or with the same tool and arguments; neither is defined for clean trials. Only missing tool differs from clean in recovery (task-clustered GEE, $p_{\text{Holm}} < .001$).
\end{table}

\subsection{Secondary analysis: asking the agent to check its work}
\label{sec:scaffold}

One line was added to the system prompt for a matched pair of models, in a second arm of 480 trials: after every tool call, briefly check whether the result is what you expected. The instruction did not move detection. For the instruct model the change was $+0.6$ points ($t(161) = 0.30$, $p = .764$) and for the reasoning variant $+2.5$ points ($t(161) = 0.94$, $p = .347$); recovery, repetition, and false alarms were likewise unchanged. Splitting by visibility, detection under quiet failures went from 60.6\% to 59.6\%, and the interaction between the instruction and visibility does not reach significance ($z = -1.35$, $p = .176$).

This arm ran on one family, so it does not license a general claim about metacognitive prompting. What it does show is that the obvious first remedy, asking for a check in words, left the gap in Figure~\ref{fig:detection} where it was. The contrast with Sun et al.\ is worth noting: there, a short disclaimer improved detection of silent errors \cite{sun2024tools}. Their models judged a single output on request, while ours had to notice the anomaly mid-task while carrying a plan, which may be where a general instruction to check loses its force.

\section{Discussion}

The results describe a specific shape of brittleness, and it is not the one an aggregate success rate would suggest.

\subsection{Agents watch the channel, not the content}

Three faults that differ in what they demand (retry, rebuild the plan, reformulate the call) produced detection rates within a point of one another, while the fault that arrives without an error string produced a rate 32 points lower. The common feature of the first three is the error itself. Reading it takes no domain knowledge and no comparison against expectation, which is what noticing a corrupted value requires: the fuel level exceeds the tank, an item has gone missing from a list that was longer a moment ago, a symbol does not match the company that was asked about. Agents do this sometimes, at rates well above the no-fault baseline, but they miss it in 41\% of trials, and in some of those they smooth the inconsistency over in their report to the user.

That is the practical warning in this study. Loud failures are the ones a monitoring system already catches, and the agent catches them too. Quiet failures are the ones that reach the user as confident output, and they are the ones agents are worst at, which inverts the usual assumption that an agent adds a layer of checking on top of the tools it calls. This agrees with the over-trust reported in single-step and multi-path settings \cite{sun2024tools, zhu2026toolmaze}; what our design adds is where in the chain it bites, at noticing rather than at acting. The human factors literature has a name for the underlying pattern, where operators accept automated output that is wrong because nothing in its presentation says otherwise \cite{mosier1999automation, parasuraman1997humans}, and the correspondence here is close enough to be worth naming even though the mechanism need not be the same.

\subsection{Acting is not the same as noticing}

The dissociation in Section IV-B is the clearest evidence that these components come apart. Reasoning variants changed their next action more often than their instruct siblings (the behaviour one would call adaptive) while treating fewer results as problems, and the two effects met in the middle at no difference in recovery. A reading consistent with the trace-level work on these models \cite{turpin2023language, chen2025reasoning} is that the deliberation phase supports generating an alternative course of action more than it supports checking a returned value against what the task implies. Whatever the cause, the practical reading is direct: the reasoning tier of a model family was not the safer choice under faults in this study, and choosing it on the assumption that more deliberation means more caution would not be supported by these data.

\subsection{Implications for deployment}

Four recommendations follow. First, monitor tool outputs rather than only tool errors: the failures agents miss are the ones your error-rate dashboards also miss, so validation belongs at the point where a value enters the agent's context. Second, do not treat detection as an alarm bell that has already rung; noticing was only modestly associated with recovery, and the association weakened once model was held fixed. Third, budget for the effort a fault costs. Recovery after a capability loss fell to 39.9\% against a 63.3\% fault-free baseline, agents returned to the same tool three or more times in up to 22\% of fault trials, and trials terminated at the step cap in nearly a third of the harder conditions, all of which turns into latency and spend in production. Fourth, judge fault effects against repeated fault-free runs: in our data, a single clean run as the reference would have made three faults look harmful that left end states within ordinary run-to-run variation.

\section{Limitations}

The boundaries of this study are worth stating plainly. The tasks are simulated environments from one benchmark family; they are executable and stateful, which is what the design needs, but they are not production services, and the fault types are a chosen four rather than an exhaustive set. Two of those four fire only when the agent calls the affected tool, at 71\% and 74\% of trials, and although drawing the target from the ground-truth path removed most of the earlier bias, the analysed subset for those conditions is still the set of trials where the agent's own choices met the fault. Detection is coded from what the agent wrote, so an agent that noticed something and said nothing counts as not noticing; the binary coding reached $\kappa = .69$ against hand-coding, which supports the comparisons made here but is not perfect agreement, and multiple human coders or behavioral signals of detection that do not rely on language would sharpen this measure. Trials were capped at 15 tool calls, and 27\% ended at that cap, so recovery rates are recovery within a bounded effort budget. Recovery is agreement with a single independent fault-free run; because two such runs agree only 63.3\% of the time, the measure is noisy, and fault effects are judged against that baseline rather than against perfect agreement. Faults fired earlier than a midpoint design would place them, and timing differed by fault type, although timeouts and corruption, the cleanest loud-quiet contrast, fired at the same median step. The reasoning comparison rests on three matched pairs, and in one of them the reasoning model wrote tool calls as text rather than emitting them, which lowers its effective action rate in a way specific to this interface; the detection difference is clearest for Qwen and holds for Claude only on matched pairs, and three pairs cannot establish a general property of reasoning models. Finally, the scaffolding arm covers one model family, and its null result should be read as one prompt on two models rather than as a verdict on prompting.

\section{Conclusion}

We injected typed faults into the tool layer of an executable agent benchmark and measured what agents do next, separating whether they notice a failure from whether they act on it and whether they recover. Across 1{,}920 trials and six models, agents treated a failure as a problem in 91.3\% of trials when the tool reported an error and 58.8\% when the tool returned a well-formed wrong value, against a 26.8\% baseline for reporting problems that did not exist. Reasoning variants noticed less than their instruct siblings and replanned more, with no gain in recovery, and a prompt asking for a check after each call did not close the gap. Against the 63.3\% agreement of two fault-free runs, only a missing tool clearly lowered recovery, to 39.9\%. In line with earlier reports of over-trust in silent tool errors, the failures that agents handle well are the ones that announce themselves. The harness, the frozen task suite, and the scored dataset are released so that other fault types, environments, and models can be measured on the same footing.

\balance

\end{document}